\documentclass{article}
\PassOptionsToPackage{numbers, sort&compress}{natbib}
\usepackage[final]{neurips_2020_ml4ad}

\usepackage[T1]{fontenc}    % use 8-bit T1 fonts
\usepackage{hyperref}       % hyperlinks
\usepackage{url}            % simple URL typesetting
\usepackage{booktabs}       % professional-quality tables
\usepackage{amsfonts}       % blackboard math symbols
\usepackage{nicefrac}       % compact symbols for 1/2, etc.
\usepackage{microtype}      % microtypography
\usepackage{bm}
\usepackage{colortbl}
\usepackage{graphicx}
\usepackage{subfigure}
\usepackage{float}
\usepackage{amsmath}
\definecolor{mygray}{gray}{.9}
\definecolor{mylavender}{gray}{0.6} 

\title{Multi-modal Trajectory Prediction for Autonomous Driving with Semantic Map and Dynamic Graph Attention Network}

\author{
Bo Dong, Hao Liu, Yu Bai, Jinbiao Lin, Zhuoran Xu, Xinyu Xu, Qi Kong \\
\texttt{JD Logistics, JD.com, China}\\
\texttt{$\{$dongbo5,liuhao163,baiyu9,linjinbiao,xuzhuoran,xinyu.xu,Qi.Kong$\}$@jd.com}
}

\def\eg{\emph{e.g.}}

\begin{document}

\maketitle

\begin{abstract}
Predicting future trajectories of surrounding obstacles is a crucial task for autonomous driving cars to achieve a high degree of road safety. There are several challenges in trajectory prediction in real-world traffic scenarios, including obeying traffic rules, dealing with social interactions, handling traffic of multi-class movement, and predicting multi-modal trajectories with probability. Inspired by people's natural habit of navigating traffic with attention to their goals and surroundings, this paper presents a unique dynamic graph attention network to solve all those challenges. The network is designed to model the dynamic social interactions among agents and conform to traffic rules with a semantic map. By extending the anchor-based method to multiple types of agents, the proposed method can predict multi-modal trajectories with probabilities for multi-class movements using a single model. We validate our approach on the proprietary autonomous driving dataset for the logistic delivery scenario and two publicly available datasets. The results show that our method outperforms state-of-the-art techniques and demonstrates the potential for trajectory prediction in real-world traffic.   

\end{abstract}

%------------------------------------------------------------------------- 
\section{Introduction}
\label{sec:intro}
%\subsection{Background}
Autonomous driving is believed \cite{geiger2012we} to have a tremendous positive impact on human society. To ensure a high degree of safety even in uncertain or dynamically changing environments, an autonomous vehicle should be able to anticipate the future trajectories of the surrounding agents (\emph{e.g.} vehicles, pedestrians, and cyclists) in advance and plan a plausible path in response to the behaviour of other agents such that the probability of collision is minimized. However, the motion trajectory of the surrounding agents is often hard to predict without explicitly knowing their intention. In this case, we need to utilize other useful information to improve safety and efficacy of the planned path of the ego-vehicle, including the observed current status of notable surrounding agents, possible physically acceptable routes in the current traffic scenario, and possible interaction outcomes with their likelihoods. Unfortunately, several challenges still exist that prevents us from utilizing this information to achieve reliable trajectory prediction. In this paper, five main challenges in trajectory prediction for autonomous driving are summarized and discussed as follows:

\noindent \textbf{Considering surrounding traffic environments}. In real-world traffic scenarios, the movement of traffic must obey traffic rules, and avoid surrounding obstacles in the meantime. That useful information can be found in the high definition (HD) map.

\noindent \textbf{Dealing with social interactions}. To avoid the collision, the trend of interacting with surrounding traffic agents needs to be captured. However, interactions between different types of traffic are very different, \eg{} the interaction between pedestrians is different from the interaction between a car and a pedestrian.

\noindent \textbf{Handling traffic of multi-class movement}. The movement patterns of different types of traffic need to be considered for autonomous driving, including cars, buses, trucks, motorcycles, bicycles, and pedestrians. In this paper, those types of traffic are divided into three categories, namely vehicles (cars, buses, and trucks), cyclists (motorcycles and bicycles) and pedestrians.

\noindent \textbf{Predicting multi-modal trajectories with probability}. In reality, people may follow several plausible ways when navigating crowd and traffic. To avoid potential collisions, the most probable future movements should be considered. 

\noindent \textbf{Probability awareness}. The probability value of each possible path of surrounding obstacles is a considerable factor in the planning and control of the autonomous driving car.

State-of-the-art methods only solve some, but not all, challenges at once as shown in Table \ref{tab:challenges}. In this paper, we present a multi-modal trajectory prediction method to tackle all these challenges, which models the dynamic social interactions among agents using Graph Attention Network (GAT) \cite{velivckovic2017graph} and semantic map. The contributions of our proposed method are summarized as follows: 
%\subsection{Contributions}
%\label{sec:contributions}
\begin{itemize}
\item[$\bullet$]
The proposed method is designed to achieve multi-modal predictions with considering traffic environments, dealing with social interactions, and predicting multi-class movement patterns with probability values, simultaneously. 
\item[$\bullet$]
In the proposed Dynamic Graph Attention Network (DGAN), Dynamic Attention Zone and GAT are combined to model the intention and habit of human driving in heterogeneous traffic scenarios.
\item[$\bullet$]
To capture complex social interactions among road agents, we combine different types of information, including a semantic HD map, observed trajectories of road agents, and the current status of the traffic.
\end{itemize}
% \noindent d. We contribute a heterogeneous traffic database for the trajectory prediction purpose in autonomous driving area. It is expected that this database will serve as a comparative benchmark for future research.
\section{Related Work}
\label{sec:relatedwork}
\begin{table}[t]
    \footnotesize
    \centering
    \setlength{\abovecaptionskip}{0cm}
    \setlength{\belowcaptionskip}{0cm}
    \caption{Comparison of challenges handled in different methods in trajectory prediction.}
    \begin{tabular}{l|c|c|c|c|c}
    \toprule
     \rowcolor{mylavender}
     Methods & Traffic Environments & Social & Multi-class & Multi-modal & Probability \\
     Social LSTM \cite{SocialLstm} &  & \checkmark  &  &  &  \\
     \rowcolor{mygray} 
     Social GAN \cite{SocialGan} &  & \checkmark &  & \checkmark &  \\
     PECNet \cite{mangalam2020not} &  & \checkmark  &  & \checkmark &  \\
     \rowcolor{mygray} 
     Argoverse \cite{Argoverse} & \checkmark &  &  &  &  \\
     Trajectron++ \cite{salzmann2020trajectron} &  & \checkmark  & \checkmark  & \checkmark  &  \\
     \rowcolor{mygray} 
     Multipath \cite{Multipath}& \checkmark &   &  & \checkmark &  \checkmark \\
     %\rowcolor{mygray}  
     DGAN (ours) & \checkmark & \checkmark& \checkmark& \checkmark& \checkmark \\ 
     \bottomrule
    \end{tabular}
    \label{tab:challenges}
\end{table}
Here, we review recent literature on trajectory prediction with social interactions. 

\textbf{RNN-related methods}. The recurrent neural network (RNN) \cite{mikolov2010recurrent} and long short term memory (LSTM) \cite{hochreiter1997long} have proven to be very effective in time-related prediction tasks. To capture social interactions between pedestrians in crowds, Alexandre \emph{et al}. \cite{SocialLstm} used a social pooling layer in LSTMs to capture social interactions based on the relative distance between different pedestrians. Chandra \emph{et al}. \cite{Traphic} introduced an LSTM-CNN hybrid method with the weighted horizon and local relative interactions in heterogeneous traffic. However, those previous studies only focus on predicting future trajectories for one class, \eg{} pedestrians or vehicles.

\textbf{GAN-related methods}. As there are multiple plausible paths that people could take in the future, several methods \citep{SocialGan, Social-bigat, Social-STGCNN} were proposed using the GAN framework to generate multiple trajectories for a given input. However, to generate multiple results for one target in practice, the generative model should be executed repeatedly with a latent vector randomly sampled from $\mathcal{N}(0, 1)$ as input. Randomly initialised inputs will generate random outcomes, which may lead to large margins between the generated results and the ground truth. To cover the most likely future paths, the number of executions has to be increased. 

\textbf{Methods that encode traffic rules}. To predict trajectories that obey traffic rules, several methods used features learned from customised semantic HD map or static-scene images to encode prior knowledge on traffic rules. Chai \emph{et al.} \cite{Multipath} proposed a multipath model to predict parametric distributions of future trajectories with HD map. It regresses offsets for each predefined anchor and predicts a Gaussian Mixture Model (GMM) at each time step. Meanwhile, with a birds-eye-view (BEV) binary image, probabilities are predicted over the fixed set of $K$ predefined anchor trajectories. Cui \emph{et al}. introduced a multi-modal architecture using a raster image from an HD map with each agent's surrounding content encoded. In \cite{Argoverse}, lane sequences were extracted from rich maps as reference lines to predict cars' trajectories. Sadeghian \emph{et al.} \cite{Sophie} presented a GAN framework integrating features encoded from the static-camera frames as the traffic rule constraints using the attention mechanism. However, those works only encode car lanes without considering pedestrian crossings, cycle lanes and other static obstacles labeled in the HD map at the same time.  
% \\
% \textbf{d. Graph-related methods}\\
% Several methods were proposed to model the relative position of the involved agents as a graph, and using a graph neural network to track the interactions. In \cite{kosaraju2019social}, the authors upgraded the SoPhie framework with bicycle-GAN \cite{zhu2017toward} and GAT \cite{velivckovic2017graph} for human trajectory prediction. In \cite{Social-STGCNN}, Mohamed \emph{et al.} extended spatial graph method to a spatio-temporal graph convolutional neural network (CNN) to predict pedestrian trajectories.
% In the rest of this paper, we show that using a Multi-Layer Per- ceptron (MLP) followed by max pooling is computationally more efficient and works as well or better than the social pooling method from [1]. Lee
% \noindent b. \textbf{GAN based methods} \\{}
% % Introducing GAN
% Agrim fist proposed a GAN based method with a MLP followed by max pooling to capture the interactions between pedestrians. In the max pooling layer, position history current status.
% \\ 
% \noindent c. \textbf{Graph related methods} \\{}
% difficult to predict trajectories within different classes methods considering multiclass: TRAPHIC attention zone \\{}
% \noindent c.\textbf{Map related methods} \\{}
% MAP/mask/camera view image fixed scene image/ rastered map/ mask TRAFFIC attention zone SOCIAL GAT
% \noindent d. \textbf{Methods with Physical Constraint} \\{}
% 1
\begin{figure}[t]
\setlength{\abovecaptionskip}{0.cm}
\setlength{\belowcaptionskip}{-0.3cm}
\centering
%\subfigure{
	\begin{minipage}[t]{0.48\textwidth}
	\centering
	\includegraphics[width=0.75\textwidth]{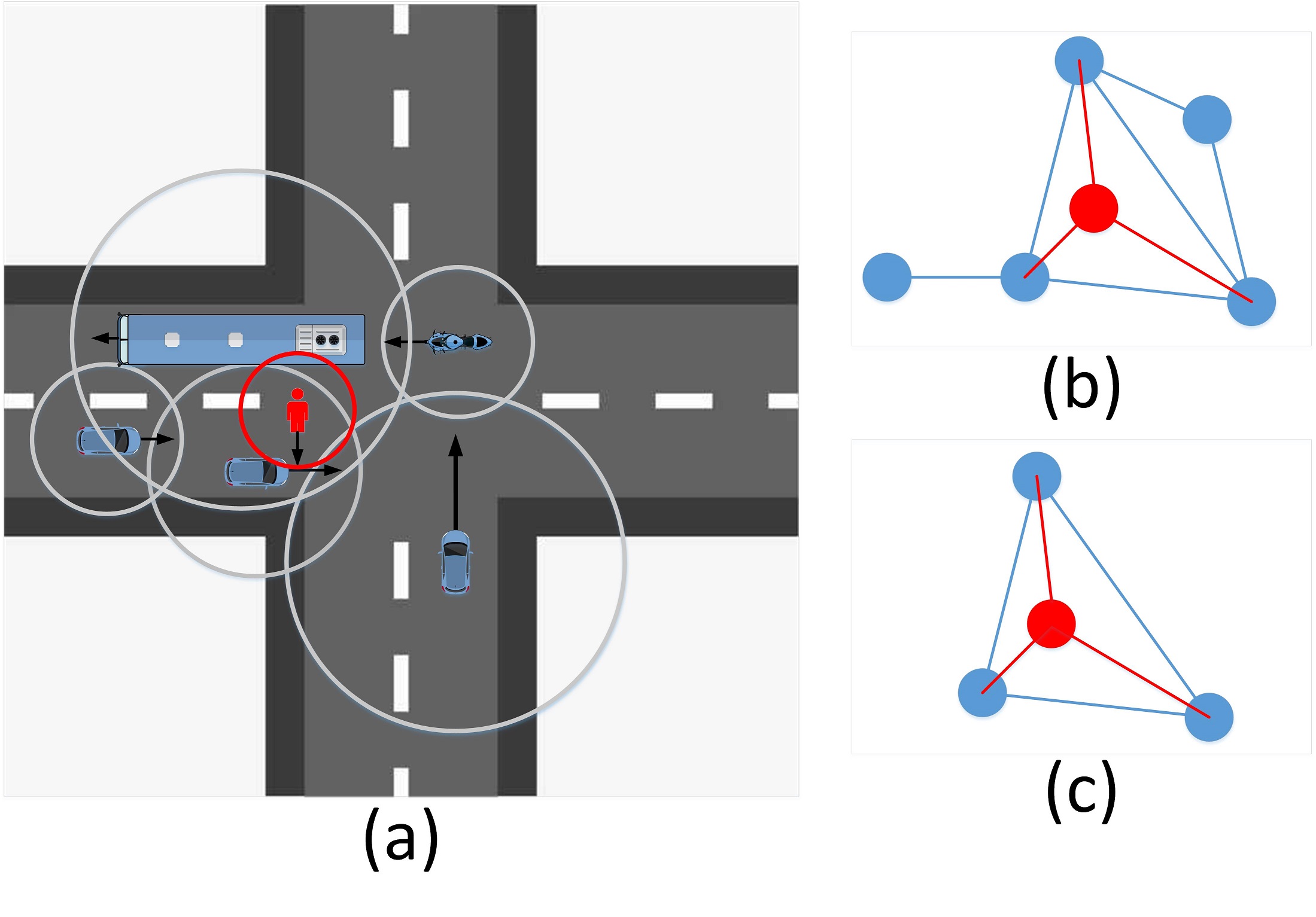}
	\caption{Dynamic attention zone and graph modelling for simulating the interaction pattern in real world traffic scenario.}
	\label{fig:1a}
	\end{minipage}
%}
\hspace{0.1in}
%\subfigure{
	\begin{minipage}[t]{0.48\textwidth}
	\centering
	\includegraphics[width=0.65\textwidth]{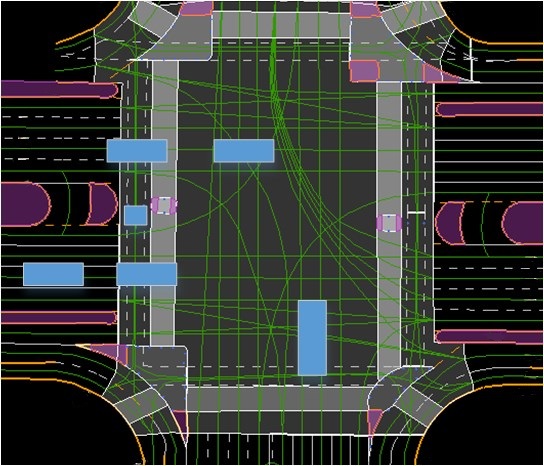}
	\caption{RGB image representation of semantic HD map for encoding the real world traffic environments.}
	\label{fig:1b}
	\end{minipage}
%}
%\label{fig:zone}
\end{figure}
\section{Methodology}
\label{sec:methodology}
\subsection{Problem Definition}
Given a set of $N$ agents in a scenario with their corresponding observed information over a time period $T_{ob}$ from time steps $1,...,t_{ob}$, our goal is to predict the future trajectories $\hat{\textbf{Y}} = \{\hat{Y_1},...,\hat{Y_N}\}$ of all agents involved in the scenario over a time period $T_{f}$ from time step $t_{ob}+1,...,t_{f}$. $N$ agents belong to multiple $c$ classes, \eg{} vehicle, cyclist, and pedestrian. Similarly, the ground truth of the future trajectory is defined as $\textbf{Y} = \{Y_1,...,Y_N\}$, where $Y_i =\{p^{t}_{i}=(x^t_i, y^t_i) | t\in\{t_{ob}+1,...,t_{f}\}$, and $i\in\{1,...,N\}$. There are three different kinds of observed information as inputs to our model, including the semantic map $map^{t_{ob}}$ of the current scenario at time stamp $t_{ob}$, the traffic state $S^{t_{ob}}_i$ of agent $i$ at current time stamp $t_{ob}$, and the observed trajectories of all agents $\textbf{X} = \{X_{1}, ..., X_{N}\}$, where $X_{i} = \{p^{t}_{i}=(x^t_i, y^t_i) | t\in\{1,...,t_{ob}\}\}$.

\subsection{Dynamic Graph Attention Network}
\label{subsec:model}
\subsubsection{Dynamic Attention Zone and Graph Modelling}
\label{subsubsec:zone}
Inspired by the real-world traffic moving pattern, a dynamic attention zone is designed to capture the normal ability of people when interacting with others in traffic. Human beings have the natural sense to choose which surrounding moving agents should be noticed by judging their current status, such as distances, headings, velocities, and sizes. Then, we model each object in the scenario to have an attention circle. Based on the intersection status of the attention circles, we can easily select surrounding agents to have social interactions with. The radius $r$ of the circle is defined as follows:
\begin{equation}
r^{t}_{i} = velocity^{t}_{i} * T_{f} + \lambda * length_{i},
\end{equation}
\noindent where $T_{f}$ represents the period of future time for prediction, and $\lambda$ is a constant value. The $velocity^{t}_{i}$ and $length_{i}$ represent the speed at time $t$ and length of object $i$, respectively. The attention zone at time $t$ covers all potential future positions over a time period $T_{f}$ based on the observed speed at the current time step and the length of the agent. If the agent accelerates or decelerates, the region of attention zone will be enlarged or reduced accordingly to predict the future movement for the next time step.  

As illustrated in Figure \ref{fig:1a}.(a), based on the current position and radius of each agent, attention zones of all agents are firstly drawn. Then, the graph of the current scenario at time step $t$ is generated based on the intersection relations of every attention zone. 

We define $G$ as $(V,E)$, in which $V = \{v_i| i\in\{1,..,N\}\}$ and $E = \{e_{ij}| \forall i,j\in\{1,..,N\}\}$, where $V$ and $E$ denotes the vertexes and edges of the graph $G$. As shown in Figure \ref{fig:1a}.(b), the graph represents the relations in the whole scenario, but in Figure \ref{fig:1a}.(c), we only focus on the partial graph related to the target in red color. The value of $e_{ij}$ will be calculated and updated in the GAT model in section \ref{subsubsec:GAT}. Each node in $V$ denotes feature embeddings calculated from three different sources including semantic map, observed trajectory, and traffic state.

\begin{figure*}[t]
\setlength{\abovecaptionskip}{0.cm}
\setlength{\belowcaptionskip}{-0.3cm}
\begin{center}
\fbox{\includegraphics[width=0.98\textwidth]{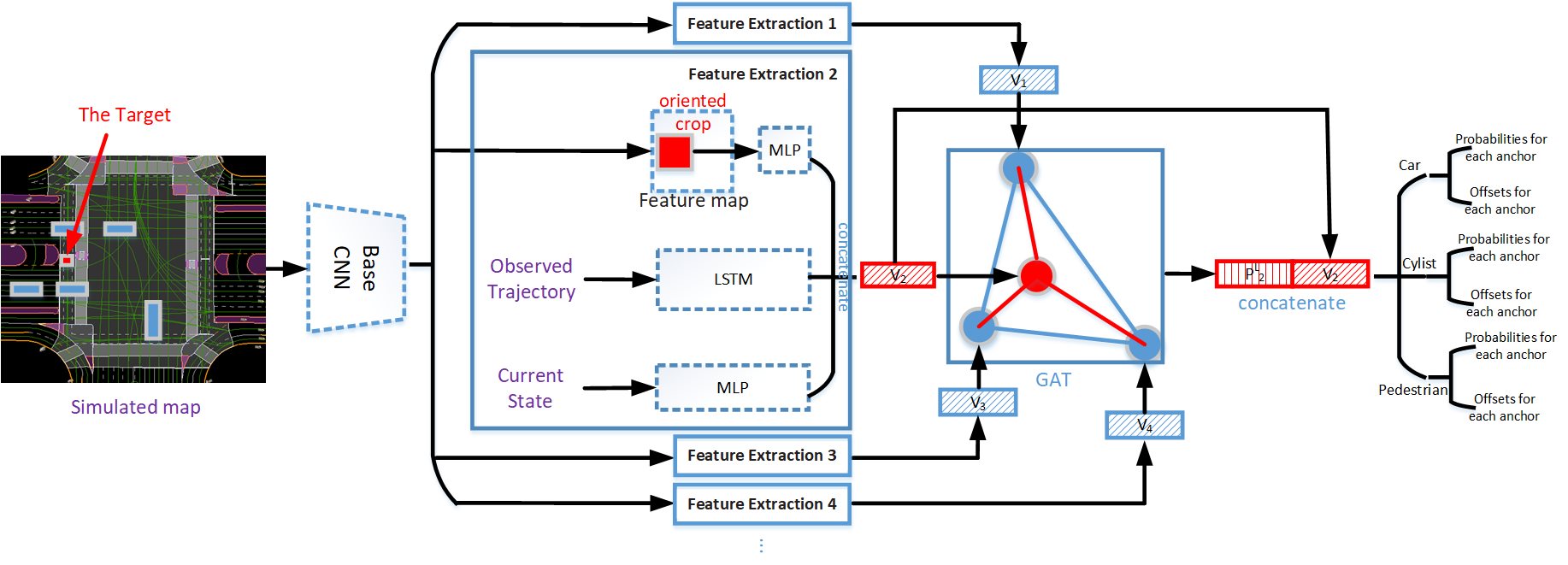}}
\end{center}
\caption{Dynamic Graph Attention Network.}
\label{fig:network}
\end{figure*}

\subsubsection{Feature Extraction}
\label{subsubsec:feature}
To make the best use of the available information, three types of features are jointly extracted from the semantic map, observed history trajectories, and current moving status.

\textbf{Semantic Map}. In autonomous driving applications, semantic HD map contains valuable traffic rule information. We create an RGB image representation to encode traffic rule information contained in semantic HD map. In the RGB image representation of the semantic HD map (Figure.\ref{fig:1b}), pink regions represent commonly seen un-movable road obstacles, \eg{} median strips or barriers. Yellow lines represent road boundaries. Grey and white regions represent pedestrian crossings and bicycle lanes. The green lines are the centre lines of lanes. Blue boxes denote movable obstacles (\emph{i.e.} it can move even though it could be stationary) in the current traffic scenario. Dotted white lines and solid white lines are the traffic lane lines and edge lines, respectively. The middle-layer output estimated by the CNN is extracted as the visual feature $V^{t_{ob}}_{map}$ to represent traffic rule information in $map^{t_{ob}}$: 
\begin{equation}
V^{t_{ob}}_{map} = CNN(map^{t_{ob}};W_{cnn}).
\end{equation}
\textbf{Observed Trajectory}. An LSTM is used to extract joint features from the observed trajectories of all involved agents. Similar to \cite{SocialGan}, we first embed the location using a single-layer multilayer perceptron (MLP) to get a fixed-length vector $e_i^t$ as the input of the LSTM cell:
\begin{equation}
\begin{split}
    e_i^t &= \phi_{ot}(X_i^t;W_{ot}),\\{}
    V^{t}_{oti} &= LSTM(V^{t-1}_{oti}, e_i^t; W_{ot}),
\end{split}
\end{equation}
\noindent where $\phi$ is an embedding function with a rectified linear unit (ReLU) nonlinearity, and $W_{ot}$ is the embedding weight. The LSTM weight ($W_{ot}$) is shared between all agents.

\textbf{traffic state}. The traffic state $S$ is very important for capturing extra information to predict the future trajectories, where $S_{i}^{t} = (velocity_{i}^{t}, acceleration_{i}^{t}, heading_{i}^{t}, width_{i}, length_{i}, c_{i})$ represent the velocity, acceleration, heading, width, length, and class of agent $i$, respectively. A simple MLP is used for encoding to get the embedding feature $V^{t}_{ts}$ of the traffic state.
\begin{equation}
    V^{t}_{tsi} = \phi_{ts}(S_{i}^{t}; W_{ts}),
\end{equation}
\noindent where $W_{ts}$ is the embedding weight of the MLP. 

The final embedding feature is defined as $V^{t_{ob}}_{i}$, which concatenates the three types of embedding calculated from the semantic map, observed trajectory, and agent status at the current time step:
\begin{equation}
    V^{t_{ob}}_{i} = concatenate(V^{t_{ob}}_{map}, V^{t_{ob}}_{oti}, V^{t_{ob}}_{tsi}).
\end{equation}
\subsubsection{Graph Attention Network}
\label{subsubsec:GAT}
The attention mechanism is found to be extremely powerful to draw global dependencies between inputs and outputs \cite{vaswani2017attention}. In attention-related methods, the GAT \cite{velivckovic2017graph} can naturally work with our proposed dynamic attention zone and graph modelling described in section \ref{subsubsec:zone}. In the graph, the vertex $V_{i}$ represents the embedding feature of agent $i$, and $e_{ij}$ represents the relative weight between an agent $i$ and its neighbour $j$ according to the graph generated from the dynamic attention zone. We use multiple stacked graph attention layers, and for each layer $l$, $W_{gat}$ is updated during training.
\begin{equation}
\begin{split}
e_{ij}&=a(W_{gat} V^{t_{ob}}_{i},W_{gat}V^{t_{ob}}_{j}),\\
a_{ij}&=softmax(e_{ij}),\\
P^{l}(i)&=\sum_{j\in N_{i}}a_{ij}W_{gat}V^{t_{ob}}_{j},
\end{split}
\end{equation}
\noindent where $e_{ij}$ indicates the importance of node $j's$ feature to node $i$, $a$ is the shared attentional mechanism described in \cite{velivckovic2017graph}, and $P^l$ is the output of the $l$th layer by summing the corresponding weighted feature of each $j$ in $N_{i}$ neighbours of agent $i$. We define $P^L$, the output from the last GAT layer $L$, as the final feature. 

Finally, the final feature $P^L$ and the original feature $V^{t_{ob}}_{i}$ are concatenated as the input of the final MLP layers $\phi_{f}$ to predict the future trajectories. We follow the idea of hierarchical classification \cite{redmon2017yolo9000} to calculate the probabilities belonging to class $c$ and anchor $k_c$.
\begin{equation}
    (prob(c)_{i}, prob(k_c|c)_{i}), \mathbf{\mu}_{ik} = \phi_{f}(concatenate(P^L, V_{i}); W_{ac}, W_{or}),
\end{equation}
\noindent where $W_{ac}$ and $W_{or}$ are weights of the MLPs for the two parallel headers, anchor classification and offset regression, respectively; $prob(c)_{i}$ and $prob(k_c|c)_{i}$ are the hierarchical probabilities for agent $i$ classified into class $c$ and anchor $k_c$; and $\mathbf{\mu}_{ik_c}$ is the predicted future trajectory offset based on the $k_c$-th anchor for the $i$-th agent. 

% 其中$P^l$是第l层的输出结果，我们把$P^L$最后一层的GAT layer（l=L）当作最后的feature

\subsection{Multi-modal Trajectory Prediction}
\label{subsec:multimodal}
The proposed method is capable of predicting multiple possible future trajectories with corresponding probability using pre-defined anchor trajectories. In this section, we present the details of multi-modal trajectory prediction.

% \subsubsection{Anchor Design and Loss Function \textcolor{red}{TODO}}
For the anchor and loss design, we follow the methods described in \cite{Multipath} and \cite{Uber-multimodal}, respectively. First, all ground-truth future trajectories are normalized in the training dataset. Then, an unsupervised classification algorithm \cite{Multipath} such as the k-means or uniform sampling algorithm, depending on datasets, is applied to obtain a fixed number of anchors with squared distance $dist(Y_{i}, Y_{j})$ between future trajectories. 
\begin{equation}
    dist(Y_{i}, Y_{j}) = \sum_{t=t_{ob}}^{t_f}||M_{i}p^t_i - M_{j}p^t_j||^2_2,
\end{equation}
\noindent where $M_i$ and $M_j$ are transform matrices which transform trajectories into the agent-centric coordinate frame with the same orientation at time step $t_{obs}$.

However, those unsupervised classification algorithms always generate redundant results for a heavily skewed distribution. In practice, we manually select anchors based on the normalized ground-truth trajectories. For each class $c$, we extract $K_c$ anchors. In total, we have $K$ anchors for anchor classification and corresponding offset regression. 

The final loss consists of anchor classification loss and trajectory offset loss:
\begin{equation}
    \mathcal{L}_{\theta} = \sum_{i=1}^{N}[\mathcal{L}^{class}_{i} + \alpha \sum_{c=1}^{C}\sum_{k_{c}=1}^{K_{c}}I_{k_{c}=k^{*}} L(\hat{Y}_{ik_{c}}, Y_{i})].
\end{equation}
$L(\hat{Y}_{ik}, Y_{i})$ represents the single-mode loss $L$ of the $i$th agent's $k_c$th anchor, where:
\begin{equation}
    L(\hat{Y}_{ik_{c}}, Y_{i}) = \frac{1}{T_{f}} \sum_{t=t_{ob}+1}^{t_{f}}\| a^{t}_{ik_{c}} + \mu_{ik_{c}}^{t}  - M_{i}p^{t}_{i}  \|_{2},
\end{equation}
\noindent where $a^{t}_{ik_{c}}$, $\mu_{ik_{c}}^{t}$, and $p^{t}_{i}$ are points at each time step $t$ of the $k_{c}$th anchor, corresponding offset based on the $k_{c}$th anchor, and $Y_{i}$, respectively.

$\mathcal{L}^{class}_{i}$ is the hierarchical classification loss \cite{redmon2017yolo9000}: 
\begin{equation}
    \mathcal{L}^{class}_{i} = -\sum_{c=1}^{C}\sum_{k_{c}=1}^{K_{c}}I_{c=c^{*}}I_{k_{c}=k_{c}^{*}}\log (prob(c)_{i}*prob(k|c)_{i}),
\end{equation}
\noindent where $I$ is the indicator function; $c^{*}$ is the ground-truth class of the agent $i$; $k_{c}^{*}$ is the index of the anchor trajectory closest to the ground-truth trajectory according to the squared distance function $dist(\hat{Y}_{ik_{c}}, Y_{i})$:
\begin{equation}
    k_{c}^* = \mathop{\arg\min}_{k_{c}\in\{1,...,K_{c}\}} dist(\hat{Y}_{ik_{c}}, Y_{i}).
\end{equation}
\section{Experiments}
\label{sec:experiments}
In this section, we evaluate the proposed methods on three datasets, including our internal proprietary logistic delivery dataset and two publicly available datasets, the Stanford drone dataset \cite{stanforddrone}, and ETC-UCY datasets. These three datasets all include trajectories of multiple agents with social interaction scenarios and birds-eye-view RGB frames used for semantic maps. 
The commonly used metrics \cite{SocialLstm, SocialGan, Traphic, Multipath}, including Average Displacement Error (ADE), Final Displacement Error (FDE), and Minimum Average Displacement Error (minADE$_{N}$), are used to assess the performances of the proposed trajectory prediction method. minADE$_{N}$ is the displacement error against the closest trajectory in the set of size $N$. minADE$_{N}$ \cite{Multipath} is computed to evaluate the method with the multi-modal property.
\begin{figure*}[t]
\setlength{\abovecaptionskip}{0.cm}
\setlength{\belowcaptionskip}{-0.5cm}
\begin{center}
\fbox{\includegraphics[width=0.98\textwidth]{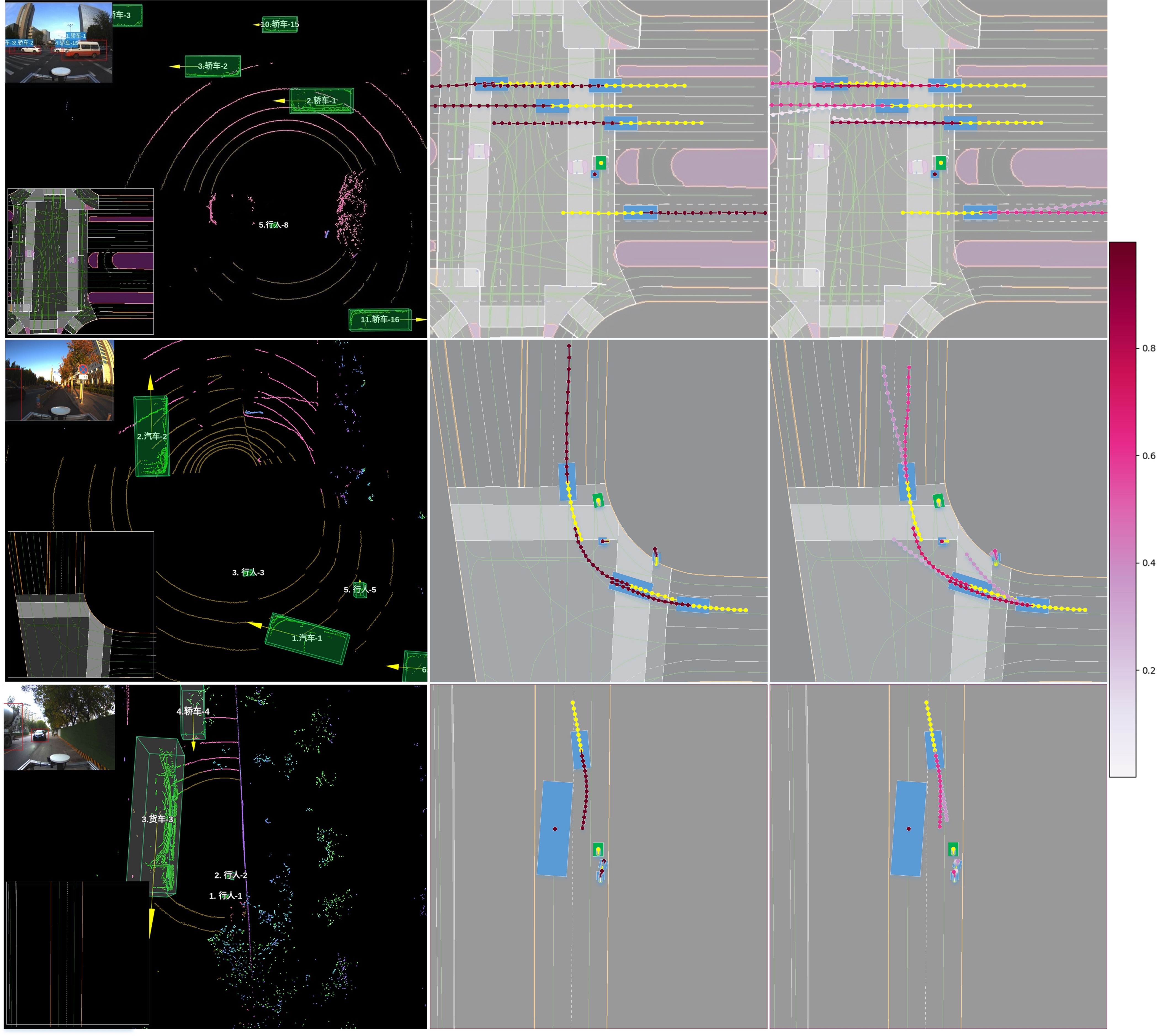}}
\end{center}
\caption{Logistic delivery dataset examples and results using our proposed method DGAN. Left: Logistic delivery dataset example, consisting of three-dimensional cloud points with manually labeled information, front camera image, and semantic map. Middle: observed in dashed yellow and future ground truth trajectories in red. Right: Prediction results using our proposed DGAN method showing up the two most likely future trajectories, and corresponding probabilities encoded in a color map to the right. The green box on the semantic map represents our autonomous driving vehicle, and only agents around it are evaluated using the proposed method.}
\label{fig:results}
\end{figure*}
\subsection{Implementation Details}
The proposed learning framework is implemented using PyTorch
Library \cite{paszke2017automatic}. For the selection of the base CNN model, we follow a similar setting as Multipath \cite{Multipath} method. Firstly, the base CNN model is a Resnet50 network with a depth multiplier of 25\%, followed by a depth-to-space operation to restore the spatial resolution of the feature map to 200$\times$200. Then we extract patches of size 11$\times$11 centered on agents locations in this feature map followed by a single-layer MLP as the representation of the traffic rules. Then, the 640-dimension feature embedding is calculated from the feature extraction block, concatenated with 256, 256 and 128-dimensional embeddings from the semantic map, observed trajectory, and current status, respectively. For the dynamic attention zone, we set the parameter $\lambda$=0.5. We train one model for each class using baseline methods, and only one model for all classes with our method.
\subsection{Logistic Delivery Dataset}
Our autonomous driving dataset for the logistic delivery purpose is collected by a vehicle equipped with multiple RGB cameras, Lidar and, radar from several regions in Beijing. We benchmark the performance of the proposed method with these baseline methods, including linear, a basic LSTM, Social LSTM(S-LSTM) \cite{SocialLstm}, Social GAN (S-GAN) \cite{SocialGan}, and Multipath \cite{Multipath}. For the logistic delivery dataset, we sample time steps every 0.2 (5Hz) from the original data and use 2 seconds of history (10 frames) to predict 3 seconds (15 frames) into the future. This dataset contains around 0.8 million agents. We extract approximately 2 million trajectories and use 90\% for training and the rest for testing.
\begin{table}[t]
    \setlength{\abovecaptionskip}{0.cm}
    \setlength{\belowcaptionskip}{0.cm}
    %\scriptsize
    \footnotesize
    \centering
    \caption{Comparison of our proposed method (DGAN) and baselines on our logistic delivery dataset. kS means the method with $K=k$ anchors using our semantic map (the S of kS stands for evaluating with semantic map).}
    \begin{tabular}{l|c|c|c|c|c|c|c|c}
    \toprule
     \rowcolor{mylavender}
     Methods  & ADEv & FDEv & ADEc & FDEc & ADEp & FDEp \\
    %  \midrule
     linear  & 3.8809 & 6.7718 & 3.7221 & 6.0352 & 1.5334 & 3.2096  \\
     \rowcolor{mygray} 
     LSTM   &3.2296 & 5.1659 & 3.0519 & 4.8564 & 1.3536 & 2.7642 \\
     S-LSTM \cite{SocialLstm} & 2.9196 & 5.0659 & 2.9519 & 4.7145 & 1.2561 & 2.6018 \\
     \rowcolor{mygray} 
     S-GAN 20VP \cite{SocialGan}  & 2.7276 & 4.5493 & 2.7567 & 4.1431 & 1.0305 & 2.2416\\
     Multipath 20S \cite{Multipath} &1.9366 & 3.2300 & 1.8573 & 2.9416 & 0.9416 & 1.8603 \\
     \rowcolor{mygray} 
     DGAN 20S (ours)  &\textbf{1.8398} & \textbf{3.0685} & \textbf{1.7593}& \textbf{2.7945} & \textbf{0.9312} & \textbf{1.8314} \\
     %\midrule
     \rowcolor{mylavender}
     Methods  & minADE$_5$v & minFDE$_5$v & minADE$_5$c & minFDE$_5$c & minADE$_5$p & minFDE$_5$p\\
    %  \midrule
     S-GAN 20VP \cite{SocialGan} & 1.6840 & 2.8835 & 1.6511 & 2.6134 & 0.6645 & 1.2848 \\
     \rowcolor{mygray} 
     %Multipath 5 \cite{Multipath}& 1.5629 & 2.8606 & 1.2250 & 2.3464 & 0.6861 & 1.1658 \\
     %Multipath 10 \cite{Multipath}& 1.5481 & 2.8358 & 1.1942 & 2.3026 & 0.5548 & 1.1603 \\
     \rowcolor{mygray} 
     %Multipath 5S \cite{Multipath}& 1.4874 & 2.5881 & 1.1575 & 2.2420 & 0.5590 & 1.1647 \\
     Multipath 20S \cite{Multipath} &  1.4595 &2.5293& 1.1391 & 2.2136 & 0.5534 & 1.1590 \\
     %\rowcolor{mygray} 
     %DGAN 10 (ours) & 1.4568 & 2.8593 & 1.2071 & 2.3138 & 0.5583& 1.1176 \\
     %DGAN 10 (ours) & 1.4519 & 2.7942 &1.1965 &2.2486 & 0.5526 &1.1047\\
     \rowcolor{mygray} 
     DGAN 20 (ours) & 1.4697 & 2.5531 & 1.1415 & 2.1918 & 0.5530 & 1.1153\\
     DGAN 20S (ours) &\textbf{1.4323} & \textbf{2.3946} & \textbf{1.1309}& \textbf{2.1636} & \textbf{0.5521} & \textbf{1.1134} \\
     \bottomrule
    \end{tabular}
    \label{tab:resultADD}
\end{table}
We compare our method on ADE, FDE, and minADE$_{5}$ against different baselines and other state-of-the-art methods. We define ADEv, FDEv, ADEc, FDEc, ADEp, and FDEp representing the ADE and FDE of vehicles, cyclists, and pedestrians, respectively. The experimental results for the logistic delivery dataset are shown in Table \ref{tab:resultADD}. As expected, the linear method performs the worst for only predicting straight paths. Our method DGAN with setting 20S (k$_c$=20 with semantic map) performs the best compared with other methods. 

Figure \ref{fig:results} illustrates the original labeled dataset, ground truth trajectories, and the top two generated results with probabilities using our method. We compare with different settings of our method, including using or not using the semantic map (Table \ref{tab:resultADD}) and the different number of K (Figure \ref{fig:anchorAnalysis}). The proposed method using the semantic map performs significantly better than without using it for the vehicle and cyclist classes. However, due to the unpredictability of movements of pedestrians and the unavailability of traffic marks in the HD map for pedestrians, the influence of the semantic map is small for the pedestrian class. The results demonstrate that our method can handle complex situations at traffic intersections. It also indicates the predicted trajectory with the maximum probability value is more likely to follow center lines of lanes guiding by the semantic map. 
\subsection{Stanford Drone Dataset}
The Stanford drone dataset \cite{stanforddrone} is collected by drones in college campus scenarios for trajectory prediction applications, consisting of birds-eye-view videos and labels of multi-class agents, including pedestrians, cyclists, and vehicles. The RGB camera frames encode traffic rule information in a semantic HD map and can serve as input to our method without any modification. For the Stanford drone dataset, we use the direction calculated from positions at the latest two observed time steps as the heading information. We use the length of the labeled bounding box as the length information of the agent. In addition to pedestrians as one class, the largest category in this database, we treat cyclists, skateboarders as one class, and the rest (carts, cars, and buses) as another class. We sample the dataset every 0.4s (2.5Hz) and use five frames of information to predict the trajectory in the next 12 frames. We evaluate the ADE, FDE, and minADE$_5$ for all agents in the test dataset compared with several state-of-the-art methods, and results are shown in Table \ref{tab:resultSDD}.  

\makeatletter\def\@captype{figure}\makeatother
\begin{minipage}[ht]{0.45\linewidth}
\setlength{\abovecaptionskip}{-0.1cm}
\centering
\includegraphics[width=0.98\textwidth]{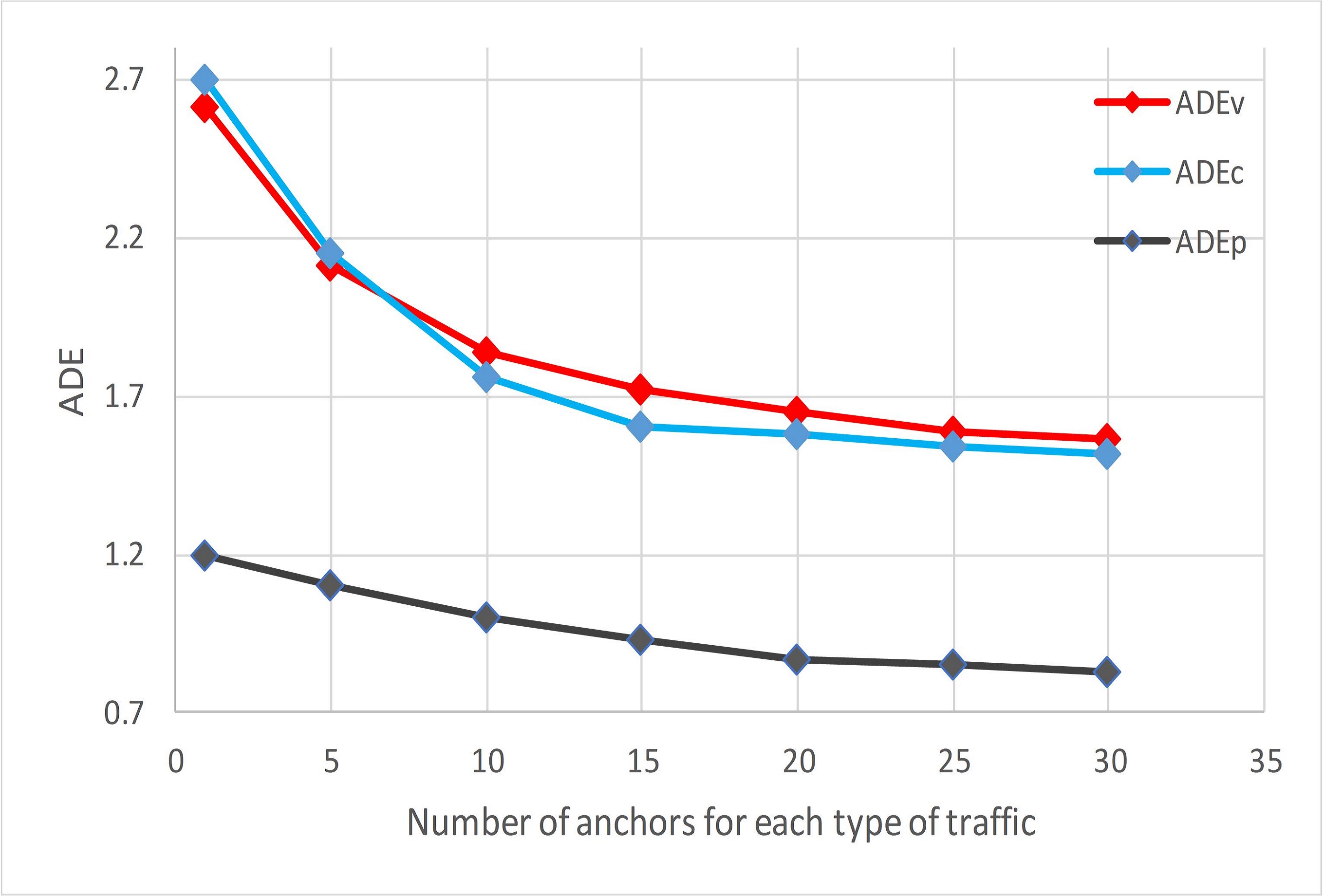}
\caption{The impact of the number of anchors K$_c$ on the final ADE result for each class.}
\label{fig:anchorAnalysis}
\end{minipage}
\hspace{0.05in}
\makeatletter\def\@captype{table}\makeatother
\begin{minipage}[ht]{0.5\linewidth}
\setlength{\abovecaptionskip}{-0.3cm}
\setlength{\belowcaptionskip}{0.1cm}
%\scriptsize
\footnotesize
\centering
\caption{Comparison of our proposed method (DGAN) and other state-of-the-art methods on the Stanford Drone Dataset. Following a similar setting with Multipath \cite{Multipath} method, distance metrics are in terms of pixels in the original resolution.}
\begin{tabular}{l|c|c|c}
\toprule
\rowcolor{mylavender}
 Methods & ADE & FDE & minADE$_5$\\
 %\midrule
 Linear    & 26.14 & 53.24 & - \\
 \rowcolor{mygray} 
 CVAE \cite{lee2017desire}     & 30.91 & 61.40 & 26.29  \\
 DESIRE-SI-IT0 \cite{lee2017desire} & 36.48 & 61.35 & 30.78\\
 \rowcolor{mygray} 
 Social Forces \cite{yamaguchi2011you} & 36.48 & 58.14 & -  \\
 S-LSTM \cite{SocialLstm}    & 31.19 & 56.97 & -\\
 \rowcolor{mygray} 
 Multipath $\mu, \Sigma$ \cite{Multipath} & 28.32  & 58.38  & 17.51 \\
 CAR-Net \cite{sadeghian2018car} & 25.72 & 51.80 & - \\
 \rowcolor{mygray} 
 DGAN (ours) & \textbf{24.53} & \textbf{50.78} & \textbf{17.28} \\
 \bottomrule
\end{tabular}
\label{tab:resultSDD}
\end{minipage}
\subsection{ETH and UCY Datasets}
The ETH\cite{pellegrini2009you} and UCY\cite{lerner2007crowds} datasets for pedestrian trajectory prediction only, include 5 scenes in total, including ETH, HOTEL, ZARA1, ZARA2, and UNIV. The trajectories were sampled every 0.4 seconds. The information in 8 frames (3.2 seconds) is observed and the model predicts the trajectories for the next 12 frames (4.8 seconds). We follow a similar setting with other relevant works \cite{SocialLstm, SocialGan} for evaluating those two datasets. Results are shown in Table \ref{tab:eth}.
\begin{table}[ht]
    \setlength{\abovecaptionskip}{-0.1cm}
    \setlength{\belowcaptionskip}{0.1cm}
    \centering
    \caption{ADE/FDE metrics for several methods on ETH and HCY datasets.}
    %\footnotesize
    \begin{tabular}{l|c|c|c|c|c|c}
    \toprule
    \rowcolor{mylavender}
    Methods & ETH & HOTEL & UNIV & ZARA1 & ZARA2 & AVG \\
    %\midrule
    Linear & 1.33/2.94 & 0.39/0.72 & 0.82/1.59 & 0.62/1.21 & 0.77/1.48 & 0.79/1.59 \\
    \rowcolor{mygray}
    LSTM & 1.09/2.41 & 0.86/1.91 & 0.61/1.31 & 0.41/0.88 & 0.52/1.11 & 0.72/1.52 \\ 
    S-LSTM \cite{SocialLstm} & 1.09/2.35  &0.79/1.76 & 0.67/1.40 & 0.47/1.00 & 0.56/1.17 & 0.72/1.54\\
    \rowcolor{mygray}
    S-GAN \cite{SocialGan} & 0.81/1.52 & 0.72/1.61 & 0.60/\textbf{1.26}& 0.34/0.69 & 0.42/0.84 & 0.58/\textbf{1.18} \\
    S-GAN-P \cite{SocialGan} & 0.87/1.62 & \textbf{0.67}/\textbf{1.37} &  0.76/1.52 & 0.35/0.68 & 0.42/0.84 & 0.61/1.21 \\
    \rowcolor{mygray}
    Ours &\textbf{0.78}/\textbf{1.50} & 0.80/1.71 & \textbf{0.59}/\textbf{1.26} & \textbf{0.31}/\textbf{0.64} & \textbf{0.39}/\textbf{0.79}  &\textbf{0.57}/\textbf{1.18} \\
    \bottomrule
    \end{tabular}
    \label{tab:eth}
\end{table}
\vspace{-0.8cm} 
\section{Conclusion}
\label{conclusion}
We have introduced a dynamic social interaction-aware model that predicts the future trajectories of agents in real-world settings to solve several challenges simultaneously. In the proposed framework, we use an encoded semantic map, the observed history trajectories, and the current status of agents as the input of the GAT. To generate the graph at the current time step, we use the dynamic attention zone to simulate the intuitive ability of people to navigate roads in real-world traffic. The proposed method is evaluated in different datasets, including our internal logistic delivery dataset and two publicly available datasets. The results demonstrate the potential ability of our method for trajectory prediction in a real-world setting. Through synthetic and real-world datasets, we have shown the benefits of the proposed method over previous methods.
\subsection{References}
\small
\label{ref}
\bibliographystyle{plain} 
\bibliography{egbib}
\end{document}